\documentclass[conference]{IEEEtran}
\IEEEoverridecommandlockouts
\usepackage{gensymb}
\usepackage[numbers]{natbib}
\usepackage{cite}
\usepackage{amsmath,amssymb,amsfonts}
\usepackage{algorithmic}
\usepackage{graphicx}
\usepackage{textcomp}
\usepackage{xcolor}
\usepackage{amsmath}
\usepackage[]{algorithm2e}
\usepackage{algorithmic}
\usepackage{tabu}

\def\BibTeX{{\rm B\kern-.05em{\sc i\kern-.025em b}\kern-.08em
    T\kern-.1667em\lower.7ex\hbox{E}\kern-.125emX}}
\begin{document}

\title{SafeNet:An Assistive Solution to Assess Incoming Threats for Premises }

\author{\IEEEauthorblockN{1\textsuperscript{st} Shahinur Alam}
\IEEEauthorblockA{\textit{Electrical and Computer Engineering} \\
\textit{The University of Memphis}\\
Memphis,TN, USA \\
salam@memphis.edu}

\and
\IEEEauthorblockN{2\textsuperscript{nd}
Md Sultan Mahmud}
\IEEEauthorblockA{\textit{Electrical and Computer Engineering} \\
\textit{The University of Memphis}\\
Memphis,TN, USA \\
mmahmud@memphis.edu}

\and
\IEEEauthorblockN{3\textsuperscript{nd} Mohammed Yeasin}
\IEEEauthorblockA{\textit{Electrical and Computer Engineering} \\
\textit{The University of Memphis}\\
Memphis,TN, USA \\
myeasin@memphis.edu}
}

\maketitle
\begin{abstract}
An assistive solution to assess incoming threats (e.g., robbery, burglary, gun violence) for homes will enhance the safety of the people with or without disabilities. This paper presents ‘SafeNet’- an integrated assistive system to generate context-oriented image descriptions to assess incoming threats. The key functionality of the system includes the detection and identification of human and generating image descriptions from the real-time video streams obtained from the cameras placed in strategic locations around the house. In this paper, we focus on developing a robust model called “SafeNet” to generate image descriptions. To interact with the system, we implemented a dialog enabled interface for creating a personalized profile from face images or videos of friends/families. To improve computational efficiency, we apply change detection to filter out frames that do not have any activity and use Faster-RCNN to detect the human presence and extract faces using Multitask Cascaded Convolutional Networks (MTCNN). Subsequently, we apply LBP/FaceNet to identify a person. SafeNet sends image descriptions to the users with an MMS containing a person’s name if any match found or as “Unknown", scene image, facial description, and contextual information. SafeNet identifies friends/families/caregiver versus intruders/unknown with an average F-score 0.97 and generates image descriptions from 10 classes with an average F-measure 0.97.
\end{abstract}


\section{Introduction}
Assessing incoming threats for homes is challenging for the people with or without disability since it requires continuous monitoring by a human observer. Sulman and colleagues \citep{sulman2008effective} found in a study that when the number of monitoring displays increases, human performance deteriorates. They reported that a human observer missed 20\% of the event while monitoring four surveillance display. However, when they increased the number of the display window to nine, missing rates rose to 60\%. The recent years have seen an upsurge of interest in developing automated systems for monitoring homes that would eliminate the necessity of human observers. The available automated commercial security solutions such as ADT \citep{ADT}, Vivint \citep{Vivint},Simplisafe \citep{Simplisafe},FrontPointSecurity \citep{FrontPointSecurity}, Honeywell \citep{Honeywell} etc. cannot detect unusual activities until an event occurs and are not designed for the people with disability. Designing assistive solutions to assess incoming threats requires consolidated knowledge about image understanding, natural language processing, and system development. In the past, a plethora of research reported on navigation \citep{gude2013blind}, expression detection \citep{anam2014expression} , currency recognition \citep{looktel}, object recognition \citep{alam2015map,kao1996object,mapelli1997role, chincha2011finding,bigham2010vizwiz} to assist people with disabilities. However, developing an automated system to assess incoming threats to homes did not receive considerable attention from the researchers. Although, the recent advancement in Machine Learning and Computer Vision, especially Convolutional Neural Network (CNN) has made the object detection \citep{krizhevsky2012imagenet,girshick2014rich, girshick2015fast,sermanet2013overfeat,he2016deep} person recognition \citep{sun2015deepid3, nezami2018face} and image captioning   task robust and efficient compared to the last decade. However, those technologies have not been used widely to build assistive solutions, especially for assessing incoming threats. To fill the void, we have developed an assistive solution to enhance the safety of people with disabilities (i.e., visually impaired, limited mobility). Our main contributions are: -1) Building a new and robust model called “SafeNet” to generate image descriptions from home monitoring cameras. 2) Collecting and processing training samples to train and evaluate SafeNet 3) Designing a recurrent neural network-based language model to generate semantically meaningful messages. Besides, we have addressed challenges related to accessibility and usability, system design, development, and integration.

\section{System Overview} 
\begin{figure*}[h]
\centerline{\includegraphics [width=\textwidth]{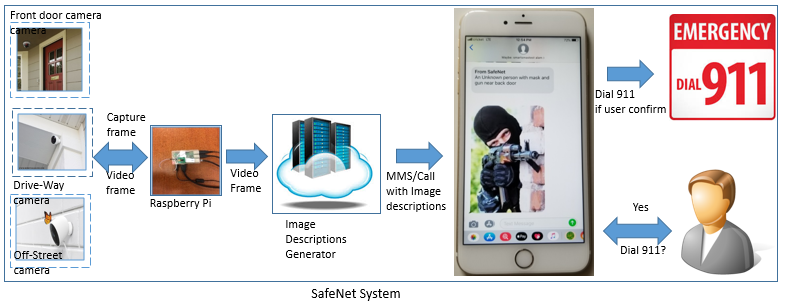}}
\caption{System architecture and process flow}
\label{architecture}
\end{figure*}

In order to use the SafeNet system, the house needs to be equipped with cameras covering the critical points such as front door, back door, driveway, off-street, etc. We assumed that people with disabilities would receive help from sighted people in installing cameras. A raspberry pi connected to the home monitoring cameras captures and sends video frames to the image descriptions generator (see “Image Description generator”). The data transmission between cameras and raspberry pi can be done using intra-home Wi-Fi or wired connection based on coverage area and distance. The image descriptions generator identifies persons from the video frames by matching faces with a personal profile and generates a short description. According to the guideline from UC Berkley police department UCPD \citep{UCPD}, a suspect can be described by information about the weapon, shirt, pants, color, eyeglasses, hair, facial hair, etc. To limit the scope of work, we have included identified person's name if any match found, location around house, whether he/she has a gun at hand or talking over the phone or wearing masks and information about facial features such as beard, mustache, eyeglasses, bald head, in the image descriptions. Here are two sample image descriptions: 1) "John at the entrance talking over the phone"; 2) "An unknown person with a gun who has beard, mustache, hair, and no-eyeglass at the back door." The system sends the image descriptions and a scene image via Multimedia Messaging (MMS) to the listed users. People with visual impairment can use the smartphone's screen reader to read out notifications. However, to make it more convenient for users’ system makes a phone call and reads out the generated image descriptions. Besides, the system records the videos and image descriptions in the persistent storage based on user preferences and allows them to query summarized history. The end-to-end process is shown in figure \ref{architecture}.

\section{System Design} 

Designing an assistive system is very challenging because it requires a comprehensive understanding about user's needs depending on the severity and type of disability, their technical adaptability, cognitive ability, and financial affordability. To design an effective system, we included a visually impaired individual who lost his vision after surgery in the development loop. The functional requirements of the system were collected through participatory design with a group of sighted and visually impaired individuals. Then, we refined and validated collected requirements by conducting a survey via Amazon Mechanical Turk \citep{Amazon} with a set of questions (see survey questions) where 30 people with disabilities participated. To address accessibility and usability issues we incorporated design thinking concept and followed “Adding Accessibility Features to Apps” (Google, 2017) technical guidelines to design a user interface.  SafeNet system has been compartmentalized into three modules: - 1. Personal Profile Module, 2. Image Description Generator Module, 3. Feedbacks Module.

\subsection{Personal Profile:} To identify a person, first, the recognition model needs to be trained with face images of friends/families. We have developed a smartphone app with four utility options (Add Person, Add Views, Delete Person, and Readout Summary) to enlist a person to the profile with voice-over interaction. The personal profile contains demographic information (Name, Email, Contact) and face images of friends/families with various expression (Joy, Sad, Surprise, Fear, Contempt, Disgust), orientation and poses so that system can recognize them robustly from different view angle, position, and distances. The app allows the user to collect face images from the photo gallery/video clip/camera preview. In camera preview mode, when a subject/friend/family stands in front of the camera, the system automatically detects face and read out the position to make sure the collected images do not have a cropped face. To capture face images from different view system guides user to rotate smartphone around the face from left to right or right to left. The captured pictures become blurry when the rotational speed is high. In order to prevent it, we provide a feedback “too fast” when the rotational speed exceeds 20 degrees per second. The collected images are sent to a Deep Webservice which is responsible for the training/re-training recognition model, versioning trained model, and data.  We have developed this \citep{RESTWeb} Deep Webservice to handle query requests robustly and seamlessly.

\subsection{Image Description Generator:} 
\begin{figure*}[h]
\centerline{\includegraphics [width=\textwidth]{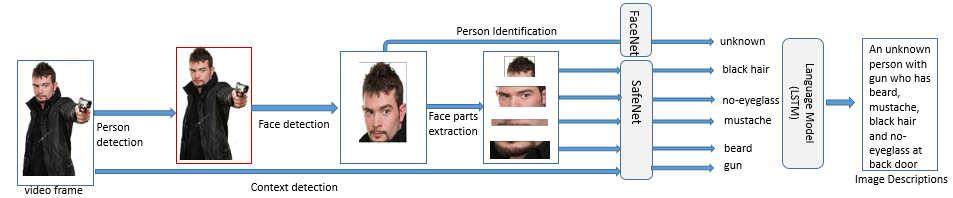}}
\caption{Workflow for generating image descriptions}
\label{workflow}
\end{figure*}

The system generates image descriptions based on the detection and recognition outcomes. The process flow of generating image descriptions is shown in figure \ref{workflow}. First, we use Faster-RCNN \citep{ren2015faster}) to detect the human presence and extract faces using Multitask Cascaded Convolutional Networks (MTCNN) \citep{zhang2016joint}. The reason for including a person detection model (Faster R-CNN) is to notify users about human presence even if there is no face found, especially when the front view of a person is not visible to the cameras.  Second, we use LBP \citep{ojala2002multiresolution} /FaceNet \citep{schroff2015facenet} to identify a person and groups by matching extracted faces with the profile (see Personal Profile). Third, face parts are extracted from detected faces using algorithm \ref{alg:crop_face}. Fourth, a facial and contextual description is generated by classifying individual face parts and images using SafeNet (see “Model Development”). The color of the hair has been determined by calculating the intensity histogram from the cropped head patch. The location of the detected person is obtained based on the source camera of the video frames. Finally, a semantically and syntactically meaningful image description is generated from obtained words by applying rule-based grammar to create initial sequences first and then refined with Long-Short Term Memory (LSTM) \citep{LSTM} language model trained with possible sequences.

\begin{algorithm}[h]
\begin{algorithmic}
\caption{Crop face Patches}
\label{alg:crop_face} 
\REQUIRE \textbf{Input: } $face\_image$
\STATE \textbf{Output: } $face\_patches$
\STATE $l  \leftarrow find\_face\_landmarks(face\_image)$
\STATE $(x, y, w, h) \leftarrow find\_face\_bound\_rect(face\_image)$ 
\STATE $(x1,y1) \leftarrow l[0]$
\STATE $(x2,y2) \leftarrow l[16]$
\STATE $(x3,y3) \leftarrow l[29]$
\STATE $(x4,y4)\leftarrow l[19]$
\STATE $(x5,y5)\leftarrow l[24]$
\STATE $(x6,y6)\leftarrow l[4]$
\STATE $(x7,y7)\leftarrow l[12]$
\STATE $(x8,y8)\leftarrow l[8]$
\STATE $(x9,y9)\leftarrow l[30]$
\STATE $(x10,y10)\leftarrow l[33]$
\STATE $(x11,y11)\leftarrow l[31]$
\STATE $(x12,y12)\leftarrow l[57]$
\STATE $ offset\_up \leftarrow integer((y8-y1)/3) $
\STATE $offset\_down \leftarrow integer((y11-y9)/3) $
\STATE $offset\_left \leftarrow integer((x1-x)/2) $
\STATE $ep \leftarrow face\_image[y1-offset\_up:y6+offset\_down,x1-offset\_left:x2+offset\_left]$
\STATE $ offset \leftarrow y8-y1 $
\IF {$y-offset  \leq  0 $}
\STATE $y \leftarrow 0$
\ELSE
\STATE $y \leftarrow y-offset$
\ENDIF
\STATE $hp \leftarrow face\_image[y:y5,x:x2]$
\STATE $bp \leftarrow face\_image[y6:y8,x6:x7]$ 
\STATE $mp \leftarrow face\_image[y10:y6,x6:x7]$
\STATE $face\_patches \leftarrow (ep,hp,bp,mp)$    
\STATE  $return face\_patches$ 
\end{algorithmic}
\end{algorithm}

\subsection{Feedback Module:} The primary task of the feedback module is sending notifications to users with image descriptions. Although smartphones are easily accessible nowadays, lots of people with a disability do not know how to utilize accessibility features such as TalkBack, Siri, etc. properly. Hence, designing an effective feedback system for people with disabilities is very challenging. Considering the technical adaptability of the users, we have included four types of feedback modes such as MMS, alert message, email, and phone call. The feedback mode can be set based on user choice. We have developed a communication API using SMTP (Simple Mail Transfer Protocol) server to send feedback messages to the users via their phone operator. To make a phone call, we are using Twilio \citep{twilio}) 3rd party service.

\section{Model development}
We have built a Convolutional Neural Network (CNN) based robust model called “SafeNet” to generate image descriptions. SafeNet is built with one input layer of dimension 320x256x3, 14 convolution layers, seven dense layers, and 5 MaxPooling layers. The output of each activation is normalized by a batch normalization layers. BatchNormalization\citep{ioffe2015batch} helps to prevent covariance shift and model overfitting. Since pooling layers reduce network dimensions very cheaply by discarding lots of spatial information, we used only five MaxPooling layers, which help to reduce the number of parameters and required computational resources. The network architecture is shown in figure \\ref{model}, and the loss curve in figure \\ref{loss}. We have come up with this architecture considering some key factors such as the size of the training dataset, overfitting vs. an underfitting problem with network depth and complexity, co-adaptation, feature learning, and predictive ability of individual neurons. The standard network such as VGG16 \citep{simonyan2014very}, ResNet50 \citep{he2016deep}, MobileNet \citep{howard2017mobilenets} etc. do not perform well with/without transfer learning (see Quantitative Evaluations) for this dataset because simple networks do not learn all distinguishing features and very complex network suffers from over-fitting problem.

\begin{figure}[h]
\centerline{\includegraphics [width=\columnwidth]{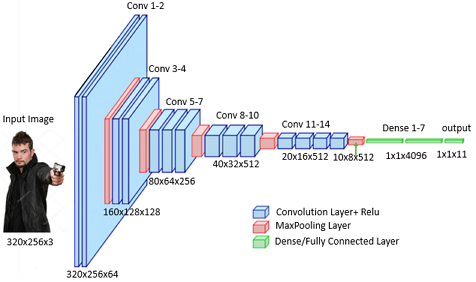}}
\caption{SafeNet network architecture}
\label{model}
\end{figure}

\begin{figure}[h]
\centerline{\includegraphics [width=\columnwidth]{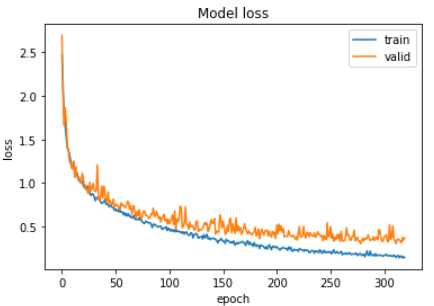}}
\caption{SafeNet network Loss Curve}
\label{loss}
\end{figure}

SafeNet has been trained for 320 epochs with a batch size of 24 using Stochastic gradient descent (SGD) optimizer. Finding optimal values for the key parameters (learning rate, momentum) of the SGD optimizer manually/iteratively is exhaustive. In addition, a low learning rate may cause slow convergence, while a high learning rate may miss the global optimum. To address this issue, Bayesian Optimization [10], a method to search hyperparameters for finding the optimal value of an unknown function has been used. The optimal values obtained from Bayesian optimizer are 0.00101 and 0.7605 respectively from a wide search space of learning rate [0.00001, 0.1] and momentum [0.5, 0.9]. The SafeNet network learned total 285,634,121 parameters. There was some criticism in the past regarding the feature learning and their interpretability in the neural network. It is indispensable to have a clear understanding of why any model performs so well, what kind of features that model learned, which part of an image played a significant role in the final classification, whether two independent neurons learned different features and all neurons have the predictive capacity, etc. In order to understand and explain those scenarios, we have visualized filters, weights, and activity of the network layer by layer (see figure \\ref{activation}.) in input pixels space. We can see from figure 5 that the bottom-layers learned edges, blobs, and textures, while upper-level layers learned higher-level abstract. All layers learned distinct features and contributed to the final classification.  

\begin{figure}[h]
\centerline{\includegraphics [width=\columnwidth]{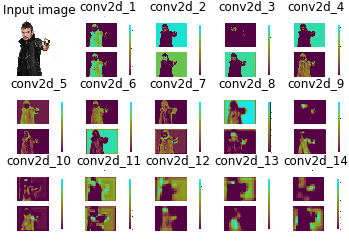}}
\caption{Two activation map/filters layer by layer}
\label{activation}
\end{figure}

\section{Data collection and Model Training}
We have included ten classes (cellphone, gun, eyeglass, mask, beard, nobeard, mustache, nomustache, baldhead, hair) to generate image descriptions and collected 8128 image samples from ImageNet \citep{deng2009imagenet},\citep{lai2011large} RGB-D  and web. Then the collected images are sorted out based on the availability of faces with/without a beard, eyeglasses, mustache, and hair. We have developed a simple and computationally efficient algorithm (see figure \\ref{facepart}) to crop different parts of a face from collected images by finding facial landmarks \citep{kazemi2014one} (see figure 6.a) and grouped similar parts together (see two sample groups in figure 6.b). Then, three-domain experts examined every single face-part and filtered out parts that are too small (less than 20x20). To make the model affine (rotation, translation, shear, scale) invariant within a certain range, we have applied data augmentations so that it can recognize face parts robustly with various orientation and head poses. Data augmentation is a very useful technique to significantly increase the diversity of the data by padding, flipping, rotating, scaling, etc. We have generated a total of 50112 samples for training and validation. Person detection and recognition model has been trained with PASCAL-VOC2012 \citep{everingham2015pascal} dataset and a profile with 180 images captured from 16 people.

\begin{figure}[h]
\centerline{\includegraphics [width=\columnwidth]{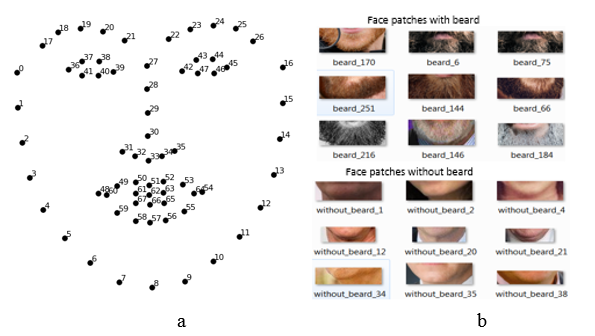}}
\caption{a) facial landmarks b) cropped samples from two groups/class.}
\label{facepart}
\end{figure}

\section{Results}

\begin{figure}[h]
\centerline{\includegraphics [width=\columnwidth]{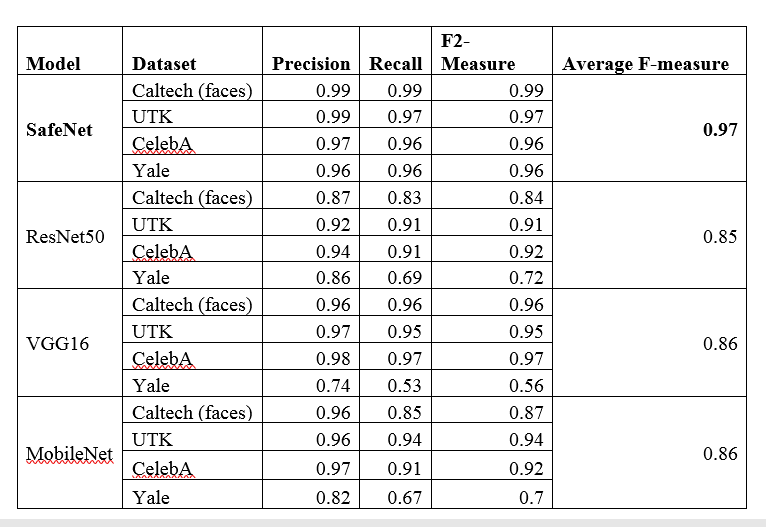}}
\caption{Average F-measure of generating image descriptions from 10 classes. }
\label{table1}
\end{figure}

The person detection and recognition model has been evaluated thoroughly with real-time video captured using Logitech C270 HD webcam placed in-front of the door from the different time points of a day to check how the system performs at different lighting conditions. The average F-measures of person identification is 0.97. SafeNet model has been evaluated with four standard datasets; Caltech \citep{Caltech}, UTKFace \citep{UTKFace}, \citep{CelebA} , Yale \citep{yagcioglu2015distributed} and image samples collected from the web, which contains people with guns, mask, and cellphone. SafeNet generates image descriptions from 10 classes with an average F2-measure 0.97, which outperformed VGG16 \citep{simonyan2014very}, ResNet50 \citep{he2016deep} and MobileNet \citep{howard2017mobilenets} for this dataset (see figure \ref{table1})

\section{Discussion}
We started developing this system with participatory design and conducted a survey with a set of 15 questions. Among them, twelve questions were asked to refine the system's functional needs and to find out user’s preferences for different functional modes (feedback modes, user interaction modes with a smartphone). Thirty people (26 males, four females) with disabilities participated in that survey where five persons were paralyzed/partially paralyzed, and eight were visually impaired, six persons with hearing disability, and 12 persons had other disabilities. We asked seven questions to understand the effectiveness and impact of our work and survey results showed (see figure 8) that our system will increase the safety and comfort of the people with disabilities. In response to a question, “how secure do you feel at home without any smart system to assess incoming threats," a visually impaired participant, who was an instructor for “technology \& apps-uses” at Clovernook Center for the Blind and Visually Impaired, Memphis, TN, said: 
\\
“Nowadays, I am just afraid of coming to the door and opening it without knowing who is there. This system will increase my comfort if I can know who is entering my house. Moreover, people will be able to replace their 250\$ doorbell and intercom system if you choose a camera that has both audio input-output and find a way to talk to the incoming person on doorstep. People like me, who is blind and retired depends on SSI (Social Security Income), cannot afford 250\$ doorbell. You wouldn’t believe how many blind people like me would buy your system if it cost 150\$ or less." 
\\
The crowd survey revealed that voice-over interaction with a smartphone is preferable than the touch screen. In addition, most of the participants selected "alert message" as a primary feedback mode since it draws more attention compare to MMS/text. Another visually impaired individual suggested us to talk to vendors (Apple/Google) so that we can bypass phone’s “Do not Disturb” list for this emergency alert message. He commented:
\\
 	"In the daytime, I want to receive all alert messages, but in the nighttime, I just want to know about the alert that saves my life."
\\

\begin{figure*}[h]
\centerline{\includegraphics [width=\textwidth]{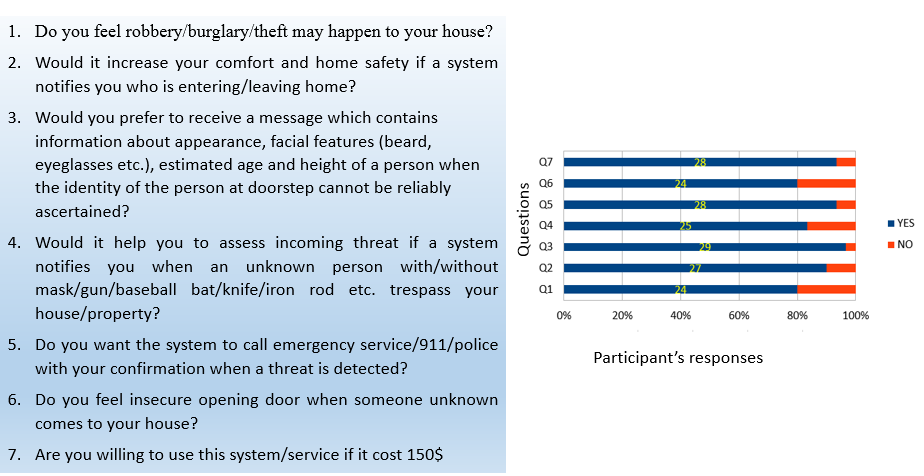}}
\caption{A user survey from 30 participants to find the effectiveness and impact of our system}
\label{survey}
\end{figure*}

\section{Conclusion}

In this paper, we have presented an interactive assistive solution to assess incoming threats which increase the safety of the people with disability. We have built a novel and robust model to generate image descriptions and collaborating with a group of people with disabilities to improve the efficacy of the system. The extensive quantitative evaluations and initial user study demonstrated that the SafeNet system enhances the safety of people with disabilities. In the alpha version, the descriptions of images have been generated from 10 categories. The visually impaired participants suggested adding more details (such as the color of shirt/pant, estimated age, and height) about a detected or recognized person in the image descriptions. Moreover, they want the system to detect a few more harmful items such as the knife, baseball bat, and iron rod.  In the beta version, we are including new features based on the user's recommendation and training SafeNet model with new items to recognize more harmful materials.

\bibliographystyle{IEEEtran}
\bibliography{SafeNet}

\end{document}